\documentclass{article}

\usepackage{graphicx} 
\usepackage{float}
\usepackage{authblk}

\usepackage{url}
\usepackage{xcolor}
\usepackage{booktabs}
\usepackage{amssymb,latexsym,amsfonts}
\usepackage{multirow}
\usepackage{tabularx}
\usepackage{color}

\usepackage[colorlinks=true, urlcolor=blue, linkcolor=red]{hyperref}
\title{Exploring the Trade-Offs: Unified Large Language Models vs Local Fine-Tuned Models for Highly-Specific Radiology NLI Task}
\newcommand*\samethanks[1][\value{footnote}]{\footnotemark[#1]}
\author[1]{Zihao Wu \thanks{These authors are equally contributing.}}
\author[2]{Lu Zhang \samethanks}
\author[2]{Chao Cao \samethanks}
\author[2]{Xiaowei Yu}
\author[1]{Haixing Dai}
\author[3]{Chong Ma}
\author[1]{Zhengliang Liu}
\author[1]{Lin Zhao}
\author[4]{Gang Li}
\author[5]{Wei Liu}
\author[6]{Quanzheng Li}
\author[7,8,9]{Dinggang Shen}
\author[6]{Xiang Li}
\author[2]{Dajiang Zhu \thanks{Co-corresponding authors: dajiang.zhu@uta.edu; tianming.liu@gmail.com}}
\author[1]{Tianming Liu \samethanks}
\affil[1]{School of Computing, The University of Georgia, Athens 30602, USA}
\affil[2]{Department of Computer Science and Engineering, The University of Texas at Arlington, Arlington 76019, USA}
\affil[3]{School of Automation, Northwestern Polytechnical University, Xi'an 710072, China}
\affil[4]{Department of Radiology and BRIC, University of North Carolina at Chapel Hill, Chapel Hill 27599, USA}
\affil[5]{Department of Radiation Oncology, Mayo Clinic, Phoenix 85054, USA}
\affil[6]{Department of Radiology, Massachusetts General Hospital and Harvard Medical School, Boston 02115, USA}
\affil[7]{School of Biomedical Engineering, ShanghaiTech University, Shanghai 201210, China}
\affil[8]{Shanghai United Imaging Intelligence Co., Ltd., Shanghai 200230, China}
\affil[9]{Shanghai Clinical Research and Trial Center, Shanghai, 201210, China}

\date{}

\begin{document}
\begin{sloppypar}

\maketitle

\begin{abstract}
Recently, ChatGPT and GPT-4 have emerged and gained immense global attention due to their unparalleled performance in language processing. Despite demonstrating impressive capability in various open-domain tasks, their adequacy in highly specific fields like radiology remains untested. Radiology presents unique linguistic phenomena distinct from open-domain data due to its specificity and complexity. Assessing the performance of large language models (LLMs) in such specific domains is crucial not only for a thorough evaluation of their overall performance but also for providing valuable insights into future model design directions: whether model design should be generic or domain-specific. To this end, in this study, we evaluate the performance of ChatGPT/GPT-4 on a radiology NLI task and compare it to other models fine-tuned specifically on task-related data samples. We also conduct a comprehensive investigation on ChatGPT/GPT-4’s reasoning ability by introducing varying levels of inference difficulty. Our results show that 1) GPT-4 outperforms ChatGPT in the radiology NLI task; 2) other specifically fine-tuned models require significant amounts of data samples to achieve comparable performance to ChatGPT/GPT-4. These findings demonstrate that constructing a generic model that is capable of solving various tasks across different domains is feasible.
    
\end{abstract}

\section{Introduction}
Natural Language Inference (NLI), also known as Recognizing Textual Entailment (RTE) \cite{haim2006second}, is a fundamental task in Natural Language Processing (NLP) that aims to determine the logical relationship between a ``hypothesis" and a ``premise". That is, given a premise, the goal of NLI is to classify the hypothesis into one of the three categories: entailment (if the hypothesis logically follows the premise), contradiction (if the hypothesis contradicts the premise), or neutral (if there is no clear relationship between the two). NLI is a challenging problem because the premise is often the only information available about the subject, and the system must reason based on that information alone. It requires inferring the relationship between statements and their relevance within the context. NLI has numerous applications in NLP and the real world, including machine translation, relation extraction, question answering, and summarization. More importantly, NLI is one of the most important tools for evaluating a system's reasoning ability.

Over the past few years, significant progress has been made in the study of NLI along with the construction of large-scale datasets, such as Stanford NLI (SNLI) \cite{bowman2015large}, Qusetion-answering NLI (QNLI), Multi-Genre NLI (MultiNLI) \cite{williams2017broad}, and Adversarial NLI (ANLI) \cite{nie2019adversarial}. These large-scale datasets have enabled substantial gains in performance on various NLP tasks and benchmarks by leveraging pre-training on massive corpora of text, followed by fine-tuning on specific tasks \cite{raffel2020exploring,liu2019roberta,yang2019xlnet,lan2019albert}. However, there is a significant limitation of these methods: while pre-training is task-agnostic, achieving satisfying performance on a desired task typically requires fine-tuning on thousands to hundreds of thousands of task-specific examples. In contrast, humans can generally perform a new language task from only a few examples or simple instructions, indicating the room for improvement in machine learning models in reasoning ability.

To narrow this gap, developing artificial general intelligence (AGI) system with human-level or greater intelligence and reasoning capability, capable of carrying out a diverse set of intellectual tasks remains a top priority for many researchers and organizations around the world. Recent advancements in large language models (LLMs), such as ChatGPT\footnote[1]{https://openai.com/blog/chatgpt}  and GPT-4\footnote[2]{https://openai.com/research/gpt-4}, have revolutionized the AGI field and opened up a new era. ChatGPT is a powerful language model that builds upon the success of the GPT-3 model \cite{Brown2020LM}, which was one of the first large language models in history. GPT-3 has a massive 175 billion parameters and was trained on a diverse collection of internet data, including Common Crawl\footnote[3]{http://commoncrawl.org/} and Wikipedia\footnote[4]{https://www.wikipedia.org/}. GPT-4 is the most powerful LLM to date, a successor of GPT-3. Although the technical details of GPT-4 have not been fully disclosed yet, it has already demonstrated superior performance over the GPT-3.5-based ChatGPT in various scenarios \cite{liu2023deid,zhao2023brain,lund2023chatting}. ChatGPT and GPT-4 have not only made breakthroughs in model architecture and pre-training techniques, but they have also made significant progress in aligning with human preferences through reinforcement learning from human feedback (RLHF) \cite{Christiano2017}. RLHF allows these models to learn from human feedback and adjust their behavior to better match human expectations and preferences. This not only improves the models’ performance and accuracy but also enhances their ability to interact with humans in a more natural and intuitive way. By learning from human feedback, these models become more adept at solving complex problems and making decisions, which is a crucial step towards achieving AGI \cite{ouyang2022training}. Moreover,  prompts \cite{liu2023pre} are used to provide specific instructions or cues to models, which offers an innovative way to convert the traditional input-output paradigm of the LLMs into a question-answering paradigm. With prompts, the LLMs can be directed to perform a specific task, without requiring the collection and labeling of large amounts of data for fine-tuning.

While ChatGPT/GPT-4 have shown impressive performance in multiple open-domain tasks, they have not been adequately tested in highly specific fields such as radiology. Clinical texts in the radiology domain exhibit unique linguistic phenomena that are distinct from open-domain data, owing to the specificity and complexity of this domain. The extensive use of medical terms and abbreviations, for instance, often leads to the out-of-vocabulary (OOV) problem. Evaluating the performance of LLMs in highly specific fields is essential not only for a thorough evaluation of the models' overall performance but also for gaining valuable insights into future model design directions. It raises important questions about whether model design should be generic or domain-specific, and whether training a single large model to address problems across all fields is the most effective approach. To this end, in this work we evaluate the performance of ChatGPT/GPT-4 on a radiology NLI task and compare it to other models fine-tuned specifically on task-related data samples. Moreover, we conduct a comprehensive investigation into ChatGPT/GPT-4’s reasoning ability by introducing varying levels of inference difficulty. Our experimental results indicate that 1) GPT-4 outperforms ChatGPT in the radiology NLI task; 2) other specifically fine-tuned models require a significant amount of data samples (over 1000 for ChatGPT and over 8000 for GPT-4) to achieve comparable performance to ChatGPT/GPT-4. These findings demonstrate that it is feasible to construct a generic model that is capable of solving various tasks across different domains.

\section{Related Work}

\subsection{Benchmarks in NLI}
In NLI domain there are several important benchmarks that have been driving the development of the entire field. 

\textbf{SNLI:} The SNLI (Stanford Natural Language Inference) dataset \cite{bowman2015large} is considered the golden classic of NLI benchmarking, with 570k examples. This dataset is based on image captions from the Flickr30k corpus, where the captions serve as premises for natural language inference. The hypothesis for each example was created manually by Mechanical Turk workers according to specific instructions. While SNLI is a valuable benchmark for evaluating NLI models, it has two significant drawbacks. First, the premises are limited to short photo descriptions and lack temporal reasoning, beliefs, or modalities. This limitation can be a significant constraint on the complexity and diversity of the data, which may not reflect real-world scenarios. Second, simple and short premises require simple and short hypotheses, making it easy for models to achieve human-level accuracy, which does not challenge them enough. Despite these limitations, the SNLI dataset remains a crucial benchmark for evaluating NLI models. 

\textbf{QNLI:} The QNLI dataset \cite{wang2018glue} is a natural language inference task that was derived from the Stanford Question Answering Dataset (SQuAD) \cite{rajpurkar2016squad}. In SQuAD 1.0, question-paragraph pairs are presented, where each paragraph is sourced from Wikipedia and contains an answer to the corresponding question. In QNLI, every question is combined with each sentence from the context (the Wikipedia paragraph) and filtered out to obtain sentence pairs with high lexical overlap. QNLI improves upon the original SQuAD task by eliminating the requirement for the model to select the exact answer and removing the assumption that the answer is always present in the input and that lexical overlap is a reliable cue.

\textbf{MNLI:} The MNLI (MultiNLI) dataset \cite{williams2017broad} is modeled after the schema of the SNLI dataset and can be used in conjunction with SNLI. MNLI comprises 433k examples and its premises are derived from ten different sources or genres, covering a broader range of spoken and written texts and providing a diverse range of data sources for evaluating NLI models. The hypothesis creation process involves presenting a crowd-worker with the premise and asking them to compose three new sentences that are either entailing, contradictory, or neutral. An essential aspect of the dataset is that only five of the ten genres appear in the training set, leaving the other five genres unseen for a model. This feature allows for estimating how well a model can generalize to unseen sources of text, thus providing an important measure of model robustness.

\textbf{SuperGLUE:} SuperGLUE \cite{wang2019superglue} is a collection of ten benchmarks that evaluate the performance of NLP models in three tasks: natural language inference, question answering, and coreference resolution. This benchmark aims to provide a single score that summarizes a model’s natural language understanding capabilities based on its performance in these tasks. It is an extension of the popular GLUE benchmark \cite{wang2018glue}, but with more complex tasks. SuperGLUE contains two natural language inference benchmarks: Recognizing Textual Entailment (RTE) \cite{RTE2006} and CommitmentBank (CB) \cite{Marneffe2019TheCI}. RTE datasets are from Wikipedia and news articles and labeled for two-class classification without neutral label. On the other hand, CommitmentBank is a corpus of short texts where at least one sentence contains an embedded clause. These embedded clauses are annotated with the degree to which the writer is committed to their truth. CB frames the resulting task as three-class textual entailment using examples from the Wall Street Journal, fiction from the British National Corpus, and Switchboard. Each example comprises a premise containing an embedded clause, and the corresponding hypothesis is the extraction of that clause. 

\textbf{ANLI:} After many new NLI benchmarks were proposed, researchers began to question the performance of these datasets. They believed that the robustness of many existing datasets, such as MNLI and SNLI, was not strong, and there was still a lot of room for improvement in the generalization performance of the trained models. These models did not achieve the original intention of the NLI task, which was to enable the model to truly learn the ability to understand natural language on the training set. Some researchers started with the model structure, trying to create models with stronger learning ability that would not get stuck in local features of the dataset. Other researchers believed that the model update speed had exceeded the dataset update speed, so they shifted their focus and attempted to build more powerful and challenging datasets. The ANLI (Adversarial NLI) dataset \cite{nie2019adversarial} is one representative of this effort. ANLI is a next-generation natural language inference benchmark that stands out for its unique data collection process, called Human-And-Model-in-the Loop Entailment Training (HAMLET). Unlike other datasets, ANLI employs a process that organizes data collection into rounds, each with increasing complexity. In each round, a SOTA model is trained, and a human annotator is tasked with generating a hypothesis that will fool the model into misclassifying the label. The example is then reviewed by two human verifiers before being added to the training set for the next round. With each round, the NLI model becomes stronger, and the contexts become more challenging. The ANLI benchmark offers an extension mechanism by adding new rounds to match the growing capabilities of SOTA models.

\subsection{LLMs} 
Large language models (LLMs) are a recent development in artificial intelligence that have gained significant attention due to their impressive performance in natural language processing tasks. These models are typically based on deep learning techniques such as transformer networks, and are trained on massive datasets of text using techniques such as unsupervised pre-training, fine-tuning, and reinforcement learning from human feedback (RLHF) \cite{Christiano2017}. Existing studies have shown that LLMs such as GPT-3 \cite{Brown2020LM} and InstructGPT \cite{Ouyang2022} can generate coherent and fluent text that is difficult to distinguish from human writing. LLMs have a wide range of applications, including language translation \cite{Vaswani2017Attn}, chatbots and virtual assistants \cite{Serban2017Dialogue}, and content generation \cite{Keskar2019LMContent}. Their potential impact on many industries has led to significant investment and research interest in LLMs, as they have the potential to revolutionize the field of natural language processing and human-machine interaction.

As for NLI, which aims to determine the logical relationship between pairs of sentences, LLMs have made significant strides in the field. One of the notable achievements of LLMs is their ability to surpass human performance in NLI tasks. For instance, the GPT-3 model has demonstrated a remarkable ability to perform NLI tasks, achieving state-of-the-art results on a variety of benchmark datasets. There have been several recent state-of-the-art works in NLI that utilize LLMs. One example is the T5 model \cite{Raffel2020Exploreing}, which uses a transformer-based architecture and a pre-training method that incorporates multiple language tasks. The T5 model achieved the best results on several benchmark datasets, surpassing previous state-of-the-art models. The RoBERTa model \cite{liu2019roberta}, a variant of the BERT model, has also achieved impressive results on several NLI tasks, including the MNLI and the Recognizing Textual Entailment (RTE) datasets. RoBERTa uses a larger training corpus and a more extensive range of pre-training tasks than BERT, resulting in improved performance on NLI tasks. Another noteworthy work in this field is the ERNIE \cite{Zhang2019ERNIE}, which integrates large-scale textual corpora and knowledge graphs (KGs) to take full advantage of lexical, syntactic, and knowledge information. ERNIE achieved impressive results on NLI datasets.

With the release of ChatGPT and GPT-4, LLMs have generated a lot of excitement in the AI community. These models are expected to have even greater capabilities than their predecessors and may push the boundaries of what is possible in natural language processing. The potential applications of these models are vast, ranging from improving language translation and generating more realistic virtual assistants to advancing research in fields such as linguistics and cognitive psychology. Specifically, ChatGPT is a member of the generative pre-trained transformer (GPT) family and is a sibling model to InstructGPT. Different from InstructGPT, which makes use of reinforcement learning from human feedback technique to fine-tune and learns user preferences by ranking feedback provided by humans, ChatGPT also makes use of RLHF but fouces on following instructions in a prompt and generating detailed responses. Though the core function of ChatGPT is to achieve human-like conversations, ChatGPT is more versatile and perform well in many fields, like debugging, writing, and so on.
ChatGPT is a highly successful AI chatbot. The original ChatGPT was fine-tuned on GPT-3.5, and GPT-4 was released on March 14, 2023, which can be accessed via OpenAI API. GPT-4 can not only interacts in a conversational style that aims to align with the user like ChatGPT, but also accepts both images and text as prompts, returning textual responses. Distinctions between ChatGPT and GPT-4 can be subtle, but differences will show up when the complexity of the task reaches a sufficient threshold. With the development in creativity, contextual-visual comprehension, and cross-modal abilities~\cite{liu2023deid}, GPT-4 improved the features and abilities of ChatGPT and can generate more diverse responses.  

\subsection{Prompt Engineering}
Prompt engineering has emerged as a novel approach accompanying the development of generative pretrained LLMs, focusing on the meticulous design and refinement of prompts submitted to these models. It is the most natural way to improve the quality and relevance of the generated text by guiding the LLMs to quickly adapt to the task pattern and retrieve particular pre-train knowledge to better couple with the domain-specific knowledge required by the task, therefore eliciting desired responses. In natural language inference (NLI) tasks that necessitate advanced reasoning coupled with domain knowledge, prompts typically commence with experts manually crafting task instructions utilizing domain-specific knowledge, followed by the incorporation of prompt-based reasoning strategies, such as the Chain of Thought \cite{wei2022chain}, to unleash the model's reasoning potential.

\textbf{Manually Designed Prompt}
Manually designed prompts intuitively reflect the user's requests and is one of the best prompting methods in most scenarios. By incorporating additional requirements and constraints, both implicit and explicit information can be encoded in the prompt for model completion. Furthermore, the quality of demonstrations plays a crucial role in in-context learning \cite{qiao2022reasoning}. Employing carefully selected, diverse, and highly representative demonstrations can enhance a model's performance, while biased demonstrations may significantly reduce it \cite{lu-etal-2022-fantastically}. Consequently, in high-specific fields such as radiology, manually designed prompts are indispensable, as the creation of task instructions and corresponding demonstrations necessitates domain-specific knowledge \cite{liu2023deid, dai2023chataug}.

\textbf{Chain of Thought}
The Chain of Thought (CoT) technique is an effective and efficient prompt-based reasoning strategy designed to enhance reasoning performance. It can be broadly categorized into three types: few-shot CoT, zero-shot CoT, and least-to-most prompting \cite{wei2022chain, kojima2022large, zhou2022least}. 

Wei et al. \cite{wei2022chain} introduced the initial CoT prompting work by supplying demonstrations with manually outlined reasoning steps preceding the final answer. This approach improved the accuracy on the GSM8K mathematical reasoning dataset \cite{cobbe2021training} to 60.1\%. Subsequent works \cite{wang2022self, li2022advance} further increased accuracy to 83\% by dividing single reasoning questions into multiple reasoning queries accompanied by intermediate correctness-check steps.

Zero-shot CoT \cite{kojima2022large} is a two-stage prompting method compared to few-shot CoT. It appends a meta-prompt "let's think step by step" to the initial prompt to obtain a reasoning process in the first stage. In the second stage, the generated reasoning process is combined with an additional prompt, "Therefore, the answer is," along with the original prompt, to elicit the model to generate the final answer based on the initial prompt and reasoning process. The authors evaluated zero-shot CoT in various reasoning tasks, achieving notable improvements.

Least-to-most prompting \cite{zhou2022least} is also a two-stage method that first decomposes a problem into several intermediate problems. In the second stage, subquestions are addressed sequentially, with each subquestion resolution facilitated by the answer to the previous subquestion. The final answer is derived from the complete set of subquestions and corresponding responses. This method significantly enhances model performance across various reasoning benchmarks.


\section{Dataset}
In this section, we present the details of converting the RadQA dataset \cite{soni-etal-2022-radqa} into the proposed RadQNLI dataset and provide statistical analysis of the resulting RadQNLI dataset.

\begin{figure}[h]
\begin{center}
\includegraphics[width=1.0\textwidth]{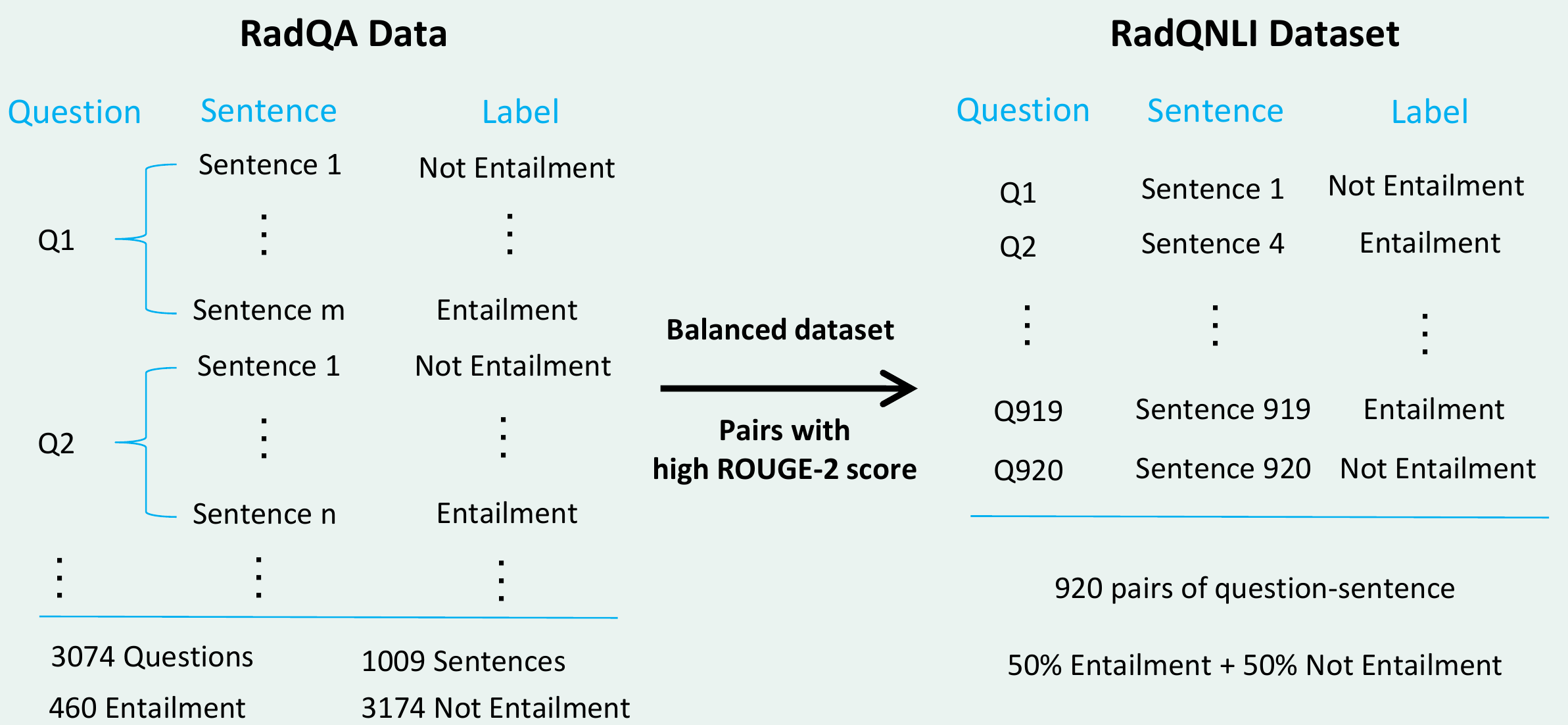}
\end{center}
\caption{The steps of creating RadQNLI dataset.} 
\label{dataset}
\end{figure}

\subsection{RadQA}
The RadQA dataset \cite{soni-etal-2022-radqa} is a radiology question-answering dataset released in 2022, containing 6,148 question-answer evidence pairs in the radiology domain. It is derived from a random sample of realistic radiology reports from the MIMIC-III database \cite{johnson2016mimic} at the patient level and then partitioned into training, validation, and test sets with a ratio of 8:1:1. In addition to the radiology reports, the dataset also includes clinical referrals authored by physicians who ordered the radiology examinations. These referrals contain actual information needs and were assembled alongside the corresponding radiology reports. To generate questions that reflect the information needed in the referrals, two annotators with medical expertise were employed. They annotated answer spans in the radiology reports based on the referrals, resulting in questions that require extensive reasoning and domain-specific knowledge to answer. In general, this dataset is well-suited for conversion into a QNLI dataset for the radiology domain.

\subsection{RadQNLI}
Drawing inspiration from previous work on transforming QA datasets into NLI datasets \cite{wang2018glue, demszky2018transforming, white-etal-2017-inference}, we convert the RadQA dataset into the proposed RadQNLI dataset in a similar way, with the goal to determine whether the context sentence contains the answer to the question. Specifically, we pair each sentence in the context with the corresponding question from that context. We assigned the labels automatically for each pair, with a label of `entailment' assigned if the context sentence contains the answer to the question and `not entailment' assigned otherwise. If the sentence contains the answer, we concatenated it with subsequent sentences to ensure the full answer span was included. To improve the dataset's quality, we introduced a selection process that uses a pre-defined ROUGE-1 score \cite{lin-2004-rouge} to screen out sentences with lower similarity (lower lexical overlap) to the questions. Our analysis shows that the ROUGE-1 score between the question and the "entailment" sentences is generally high, indicating a high level of similarity. Conversely, the ROUGE-1 score between the question and the "not entailment" sentences is typically low, indicating a lower level of similarity. A larger ROUGE-1 score will screen out more "not entailment" sentences that have a lower similarity with the question, leading to preserving fewer "not entailment" samples. This can make the NLI task more challenging as both the "entailment" and remaining "not entailment" sentences will have higher similarity to the question, making it harder for the model to distinguish between them. To ensure a suitable difficulty level and balance between the two categories, we selected a ROUGE-1 score of 0.2 as the threshold for lexical overlap between the context sentence and the question. By doing so, we aimed to maintain the challenging nature of the RadQNLI, thereby making it appropriate for training and evaluating advanced models with robust reasoning capabilities.

\begin{table}[!h]
\centering
\caption{Sample Size of RadQNLI Dataset}
{
\begin{tabular}{lcc}
\toprule
Measure & Raw & Threshold with R1$\geq$0.2\\
\midrule
Avg question tokens & 12.34 & 12.32 \\
Avg sentence tokens & 21.66 & 20.66 \\
\# of unique questions & 2771 & 2614 \\
\# of unique sentences & 11829 & 5355 \\
\# of entailment & 4391 & 4391 \\
\# of not entailment & 35504 & 5447 \\
Total samples & 39895 & 9838 \\
\bottomrule
\end{tabular}
\label{number of sample}
}
\end{table}

\begin{table}[!h]
\centering
\caption{Modalities and Organs Covered by RadQNLI Dataset}
{
\begin{tabular}{cc|cc}
\toprule
 Top 5 modalities & Percentage & Top 5 organs & Percentage\\
\midrule
X-Ray & 12.33\% & Lung/chest & 29.45\% \\
Computed Tomography (CT) & 4.61\% & Vessel & 11.26\% \\
Magnetic Resonance (MR) & 4.07\% & Brain & 9.09\% \\
CT Angiography (CTA) & 3.13\% & Abdomen & 8.94\% \\
Unknow & 77.35\% & Cardiac & 5.51\% \\
\bottomrule
\end{tabular}
}
\label{Dataset Analysis}
\end{table}

Table \ref{number of sample} and Table \ref{Dataset Analysis} presents the specifics of the newly created RadQNLI dataset, which comprises 2614 questions and 5355 sentences. From these questions and sentences, we generated 9838 question-sentence pairs, with 4391 labeled as "entailment" and 5447 labeled as "not entailment". These samples covers various modalities of imaging information for multiple organs.
 


\section{Method}
In this section, we present the methodology employed for the radiology QNLI task utilizing ChatGPT and GPT-4. We implemented ChatGPT and GPT-4 alongside various prompt designs for natural language inference and contrasted the outcomes with several baseline models.

\subsection{Zero-shot and Few-Shot Entailment Inference}
The zero-shot and few-shot in-context learning capabilities of ChatGPT and GPT-4 obviate the need for supervised fine-tuning, facilitating rapid adaptation to our novel Radiology QNLI task to evaluate the models' context awareness and understanding in the radiology domain.

\begin{figure}[h]
\begin{center}
\includegraphics[width=1.0\textwidth]{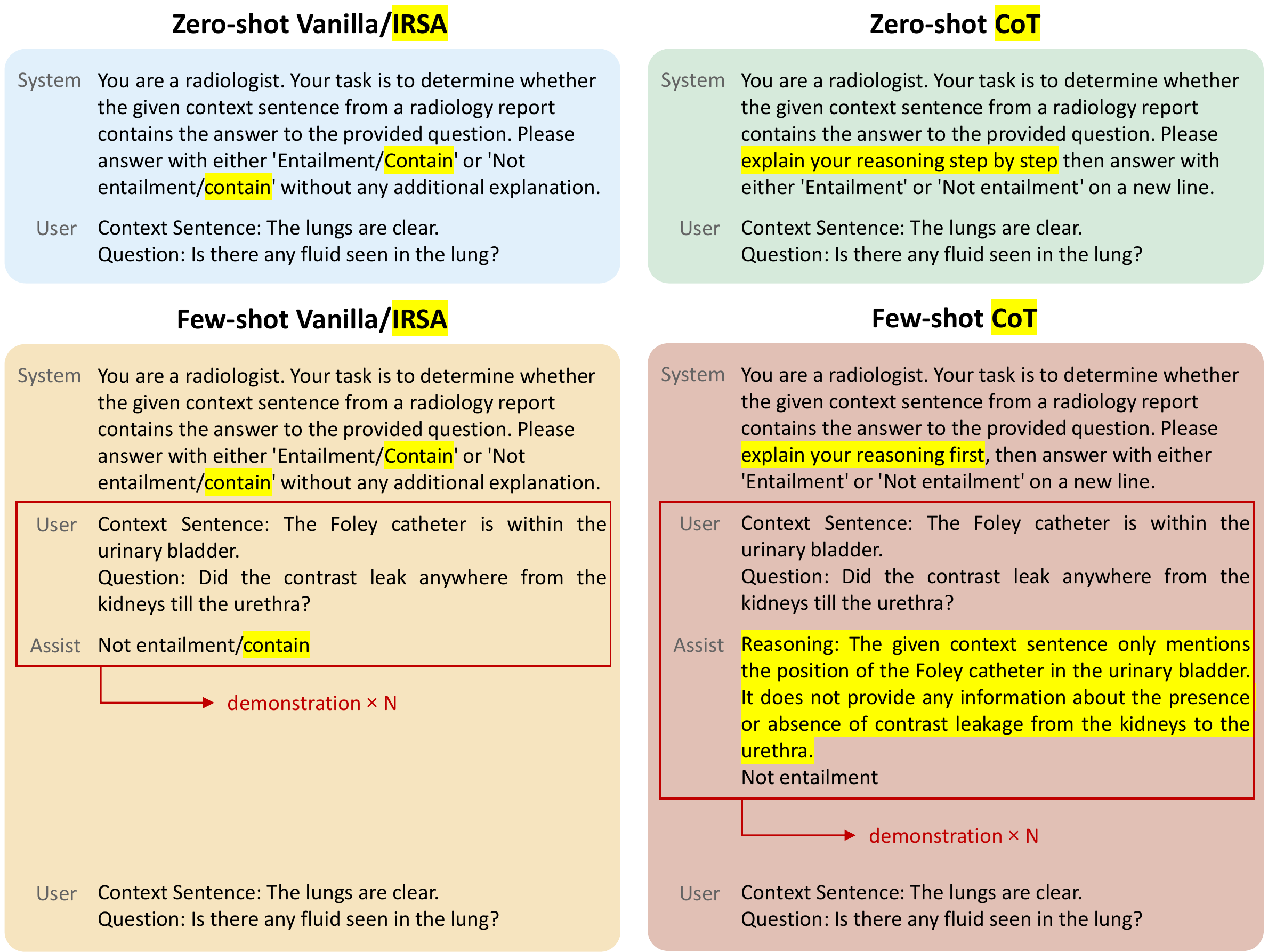}
\end{center}
\caption{Prompts used in the experiments. Modifications for IRSA and CoT are highlighted.} 
\label{Prompts}
\end{figure}

\textbf{Zero-Shot}:
We formulated the zero-shot prompt using only the task instructions and context sentence-question pairs, which directly requested ChatGPT and GPT-4 to determine if it was an entailment or not without providing any labeled examples. The prompt for zero-shot inference is presented in Fig.~\ref{Prompts}.

\textbf{Few-Shot}:
We constructed the few-shot prompt by incorporating 1) task instructions and 2) 10 context sentence-question pairs with corresponding labels as examples for in-context learning. The context-question pair under evaluation was appended at the end. The prompt for few-shot inference is presented in Fig.~\ref{Prompts}.

\subsection{Inference with Chain of Thought}
Prior research has indicated that reasoning tasks can be significantly enhanced by employing Chain of Thought (CoT) prompting, which requests the model to think step by step like a human, resolving multi-step complex reasoning questions sequentially to reduce difficulty and guide the model to follow the chain of thought to ultimately arrive at the answer. We employed the CoT scheme in both zero-shot and few-shot inference.

\textbf{Zero-Shot with CoT}:
Inspired by previous work, we included ``Please explain your reasoning first and then answer" in the task instruction prompt as presented Fig.~\ref{Prompts}.

\textbf{Few-Shot with CoT}:
To obtain the CoT for the 10 provided examples, we utilized GPT-4 to auto-generate the reasoning steps, as GPT-4 excels at tasks requiring advanced reasoning and complex instruction comprehension. In cases where GPT-4 output the incorrect label, we prompted the model to regenerate the correct one. The auto-generated reasoning was subsequently refined by a human to form the CoT. The refined CoT for each example, along with the label, was provided as the desired output. The few-shot with CoT prompt is presented in Fig.~\ref{Prompts}.

\subsection{Instruction-Response Semantic Alignment}
The original labels of our RadQNLI dataset are 'entailment' and 'not entailment', identical to the QNLI dataset. We instructed models to respond with simpler and more common words: either "contain" or "not contain", as opposed to the original label, to better align the semantics in the task instruction prompt and further reduce difficulty. After collecting the responses for each test pair, we replaced them with the original labels back for evaluation.

\subsection{Baseline Models}
To compare with ChatGPT and GPT-4, we tested the following baseline models in two categories.

\subsubsection{Text Generation Models}
For these text generation models, which possess the ability of in-context learning, we employed the same prompt as ChatGPT and GPT-4 to request models to perform inference on the test set.

\textbf{Llama:}
LLaMA (Large Language Model Meta AI) is a series of foundational large language models released by Meta AI in February 2023. Researchers trained these models, ranging from 700 million to 65 billion parameters (7B, 13B, 33B, and 65B), on trillions of tokens. The 65B model maintains a relatively smaller size while being competitive with GPT-3.5 and other top LLMs ~\cite{touvron2023llama}.

\textbf{Alpaca:}
Alpaca~\cite{alpaca} is an instruction-following model released by Stanford University. It was fine-tuned from Meta’s LLaMA 7B model and trained on 52K instruction-following demonstrations generated using text-davinci-003 dataset.Despite being a small and easily reproducible model, Alpaca performs comparably to text-davinci-003 in terms of performance.

\textbf{Bloomz:}
BLOOMZ~\cite{muennighoff2022crosslingual} is a family of models capable of following human instructions in dozens of languages zero-shot. They were generated by finetuning BLOOM~\cite{scao2022bloom} pretrained multilingual language models on crosslingual task mixture dataset(xP3)~\cite{muennighoff2022crosslingual}. Bloomz has been found the ability of crosslingual generalization and performing effectively on tasks or languages that were previously unseen.

\subsubsection{BERT-Based Fine-Tuned Models}
We utilized pretrained models, fine-tuned them on the training and validation sets, and then evaluated their performance on the test set for comparison.

\textbf{BERT:} 
In 2018, Google released Bidirectional Encoder Representations from Transformers (BERT) ~\cite{devlin2018bert}, a collection of masked-language models. BERT stands out as the first unsupervised language representation model that is deeply bidirectional, meaning it is capable of analyzing language in both directions. This capability enables BERT to extract high-quality language features that can be fine-tuned on domain-specific data for improved performance. Notably, BERT's pre-training is conducted solely on plain text corpora. 

\textbf{DistilBERT:}
DistilBERT is a compact and efficient  Transformer model based on BERT. By distilling BERT, DistilBert possesses a smaller parameter count than bert-base-uncased by 40\% and performs faster by 60\%. DistilBERT achieves comparable performance to BERT on the General Language Understanding Evaluation (GLUE) language understanding benchmark, maintaining an accuracy rate of over 95\%.

\textbf{RoBERTa:} 
The RoBERTa model~\cite{liu2019roberta} was based on BERT and share same architecture. By adding dynamic masking, removing the next-sentence pretraining objective and training with larger batches, RoBERTa is a better reimplementation of BERT with some modifications to the key hyperparameters.

\textbf{BioBERT:} 
BioBERT (Bidirectional Encoder Representations from Transformers for Biomedical Text Mining)~\cite{lee2020biobert} is a domain-specific language representation model pre-trained on large-scale biomedical corpora PubMed abstracts (PubMed) and PubMed Central full-text articles
(PMC). When compared to BERT and other pre-existing models, BioBERT demonstrates superior performance in various biomedical text mining tasks

\textbf{ClinicalBERT:} 
ClinicalBERT~\cite{alsentzer-etal-2019-publicly}, which is based on BERT, utilizes bidirectional transformers to analyze clinical notes from the Medical Information Mart for Intensive Care III (MIMIC-III) dataset. The model has been evaluated by medical professionals and has been found to establish high-quality relationships between medical concepts, as evidenced by their judgements. 

\subsection{Implementation}
The ChatGPT model used in this paper is 'gpt-3.5-turbo-0301', released on March 1st, 2023. The GPT-4 model employed in this paper is 'gpt-4-0314', released on March 14th, 2023. The baseline pretrained models and corresponding experiments were conducted using the Hugging Face Transformers package on an NVIDIA A100 GPU.

\subsection{Evaluation Metrics}
We employed accuracy as the evaluation metric to assess the models' reasoning and inference performance. All experiments involving ChatGPT and GPT-4 models report the average accuracy with the standard deviation of three repetitions to eliminate randomness and improve reproducibility.

\begin{equation}\label{eq_accuracy}
   Accuracy = \frac{TP+TN}{TP+TN+FP+FN} 
\end{equation}

\section{Experiment and Analysis}
\subsection{ChatGPT and GPT-4 with Different Prompting}
\begin{figure}[!h]
\begin{center}
\includegraphics[width=1.0\textwidth]{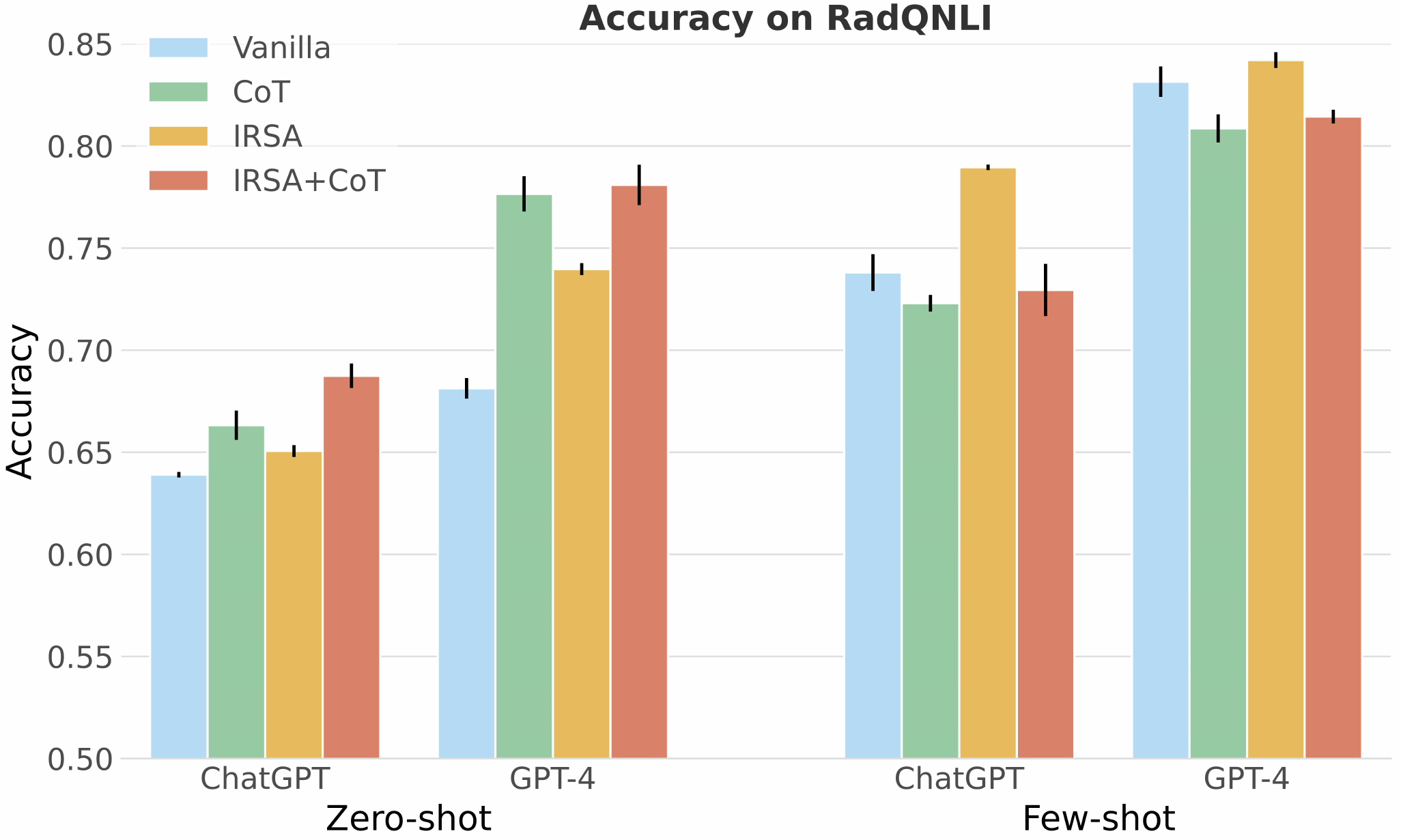}
\end{center}
\caption{We assessed the impact of various prompt methods (as detailed in Sec.~\ref{prompt methods}) on the performance of ChatGPT and GPT-4 for both zero-shot and few-shot scenarios.} 
\label{Ablation}
\end{figure}

We assess the inference performance of ChatGPT and GPT-4 using three techniques: zero-shot and few-shot in context learning, chain of thought prompting, and instruction-response semantic alignment prompting. Fig.~\ref{Ablation} displays the inference accuracy of ChatGPT and GPT-4 under various settings, with zero-shot results on the left and few-shot results on the right. In each setting, we report the outcomes for four types of prompting designs: the blue bars signify the vanilla prompt design, the green bars represent the vanilla prompt with chain of thought, the yellow bars indicate the vanilla prompt with instruction-response semantic alignment, and the red bars denote the vanilla prompt with both CoT and IRSA. Detailed prompts for the different prompting techniques can be found in Fig.~\ref{Ablation}. As depicted in Fig.~\ref{Ablation}, GPT-4 outperforms ChatGPT in all four prompting designs for both zero-shot and few-shot settings, showcasing a powerful capability for human-level understanding in language tasks \cite{openai2023gpt4}. Additionally, we provide an ablation analysis of the three techniques below.

\textbf{Zero-shot and Few-shot Prompting:}
\label{prompt methods}
Based on the results using the vanilla prompt design (blue bars), the few-shot setting considerably surpasses zero-shot prompting for both ChatGPT and GPT-4, with accuracy increasing from 0.6390 to 0.7380 and from 0.6813 to 0.8315, respectively. The improvement of over 0.10 demonstrates that few-shot prompting, by simply providing demonstrations during inference, is an efficient and effective technique for directly enhancing model performance on the QNLI task. Conversely, GPT-4 improves accuracy by 0.423 in the zero-shot setting, while in the few-shot setting, the improvement reaches 0.935, nearly doubling compared to the zero-shot setting. This significant enhancement in the few-shot setting indicates that GPT-4 can better comprehend and utilize the examples in the prompt, serving as a guide to retrieve and revise pre-trained knowledge and generate desired responses that are highly aligned and consistent with the examples.

\textbf{Chain of Thought:}
The results of experiments utilizing chain of thought prompting are represented by the green bars in Fig.~\ref{Ablation}. In the zero-shot setting, as displayed on the left side of Fig.~\ref{Ablation}, compared to the vanilla prompt design, CoT prompting enhances accuracy for both ChatGPT and GPT-4 by 0.243 and 0.952, respectively. The differing percentages confirm that GPT-4 is better facilitated by CoT in the zero-shot setting, as its exceptional reasoning ability can be effectively coupled with CoT. However, in the few-shot setting, shown on the right side of Fig.~\ref{Ablation}, there is a roughly 0.2 decrease in prediction accuracy for both ChatGPT and GPT-4. A possible explanation for the contrasting effects of CoT on zero-shot and few-shot settings is that in the zero-shot setting, without examples to follow, the model must rely solely on pre-trained knowledge to generate responses. In this scenario, CoT enables the model to break down complex tasks into a chain of smaller, more manageable tasks and solve them sequentially. This approach allows the model to accurately retrieve specific pre-trained knowledge for each smaller task and better utilize reasoning and context awareness abilities, leading to a significant improvement in accuracy. Conversely, in the few-shot setting, although the provided examples can help the model better understand complex tasks by demonstrating desired reasoning steps and corresponding responses, the model may also overfit on the examples shown in the prompt, particularly when the model is pre-set to focus on deterministic, realistic data with less creativity and randomness. Furthermore, CoT prompting in the few-shot setting significantly increases the number of tokens in the prompt, which in turn increases the complexity of the task. The model may concentrate on the examples and neglect the essential instruction obscured by the numerous examples.

\textbf{Instruction-Response semantic alignment:}
The results of experiments using Instruction-Response Semantic Alignment (IRSA) prompting are indicated by the yellow bars in Fig.~\ref{Ablation}. Compared to the vanilla prompt design, IRSA prompting outperforms in all four scenarios—zero-shot or few-shot settings, with or without IRSA—increasing prediction accuracy by an average of 0.330. When compared to CoT prompting, in the zero-shot setting, IRSA prompting achieves competitive performance. In the few-shot setting, while CoT prompting falls short, using IRSA prompting alone enables both ChatGPT and GPT-4 models to achieve their best performance at 0.7895 and 0.8420, respectively. This suggests that IRSA prompting is an effective and efficient technique for QNLI tasks, as it eliminates the additional cost of unnecessary reasoning components in responses when using paid LLMs like ChatGPT and GPT-4.

\subsection{Comparison with recently released LLMs}
We report the results of different large-scale language models on the RadQNLI dataset in Table~\ref{Other_LLMs}. The findings demonstrate that ChatGPT and GPT-4 explicitly outperform all other recently released LLMs.

\begin{table}[ht]
\centering
\caption{The accuracy(ACC) of the five recent LLMs tested on proposed RadQNLI dataset with zero-shot incontext learning. \textcolor{red}{Red} and \textcolor{blue}{blue} denotes the best and the second-best results, respectively.}
\label{Other_LLMs}
\begin{tabular}{lcccc}
\toprule
Methods & ACC\\
\midrule
Bloomz-7b &0.5109\\
Alpaca  &0.5435\\
LLaMA-7b  &0.5488\\
ChatGPT  &\textcolor{blue}{0.6875}\\
GPT-4 &\textcolor{red}{0.7809}\\
\bottomrule
\end{tabular}
\end{table}

\begin{figure}[ht]
\begin{center}
\includegraphics[width=1.0\textwidth]{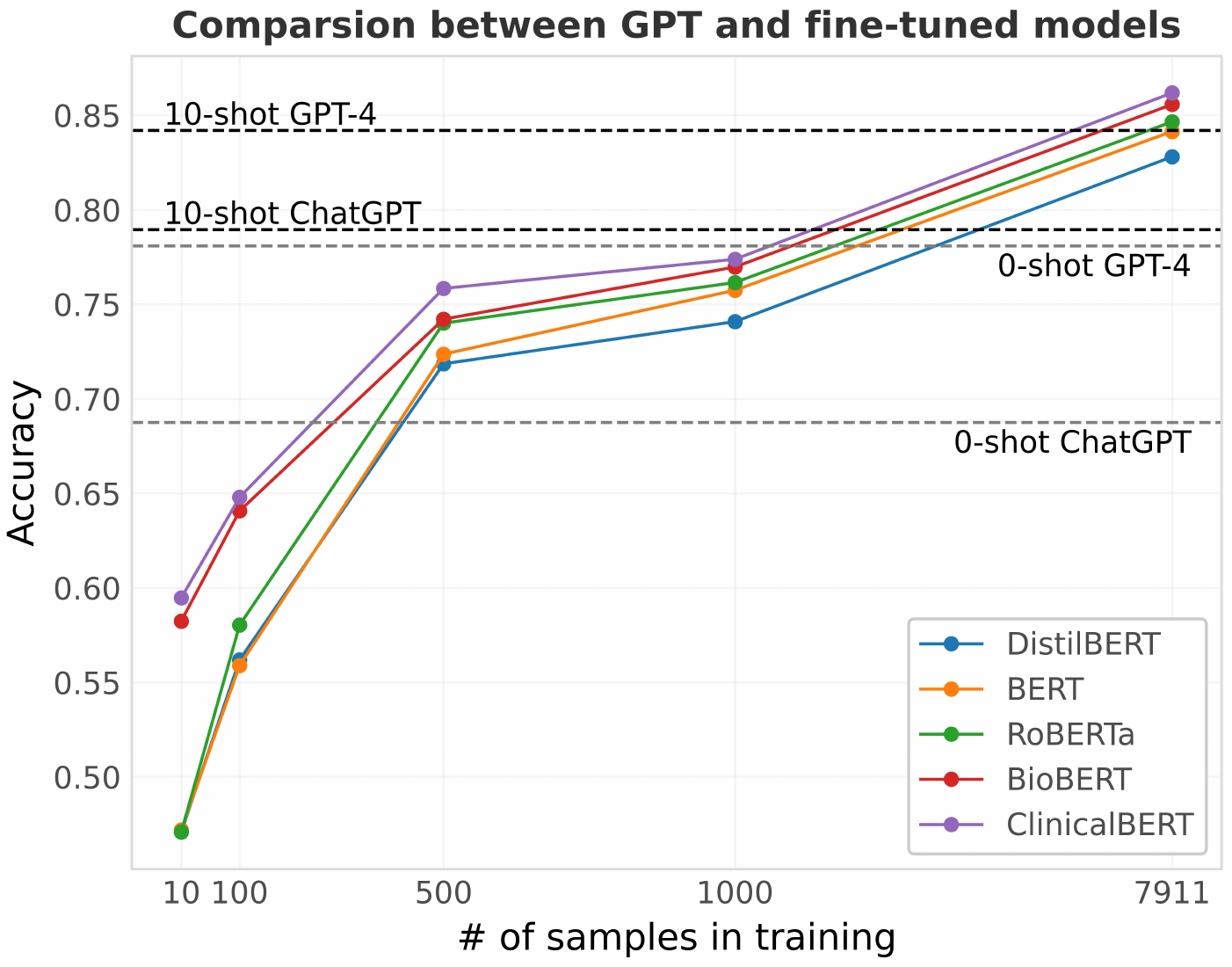}
\end{center}
\caption{We compared the performance of ChatGPT, GPT-4, and five other fine-tuned LLMs. To evaluate the impact of prompt examples on ChatGPT and GPT-4, we tested their performance with zero-shot and 10-shot examples, where greater example usage resulted in higher accuracy. Additionally, we explored the effect of increasing the size of the training dataset for the fine-tuned LLMs, which led to improved accuracy.} 
\label{GPTvsBERT}
\end{figure}

\subsection{Comparison with BERT-based Fine-tuning}
The proposed RadQNLI dataset consists of 7,911 training samples. We randomly select 10, 100, 500, 1,000, and 7,911 training samples from the original training set and conduct five fine-tuning experiments with different training set sizes to explore the effect of training set size on the performance of fine-tuned BERT-based models in QNLI tasks and the feasibility of fine-tuning BERT-based models in few-shot-like settings. The entire validation and test sets are used in these experiments. The results, along with the best performances of ChatGPT and GPT-4, are displayed in Fig.~\ref{GPTvsBERT} for comparison. With the increment of training set size, the prediction accuracy of all BERT-based models continuously improves, from an average accuracy of 0.5181 across five models with 10 training samples to 0.8467 with the full training set. Using the exact same 10 training samples as in the fine-tuning experiment as examples in few-shot prompting, ChatGPT achieves a competitive accuracy of 0.7895. GPT-4 attains an accuracy of 0.8420, very close to ClinicalBERT, the best BERT-based model fully fine-tuned on the RadQNLI dataset. These results suggest that ChatGPT and GPT-4 are more efficient than traditional BERT-based fine-tuning by utilizing significantly fewer labeled data and eliminating the fine-tuning process altogether.

\subsection{Difficulty Level of RadQNLI}
\label{Difficulty Level}
To further assess the difficulty level of the RadQNLI dataset, we experimented with different parameter settings of ROUGE-1 for thresholding lexical overlap, as described in the Dataset section. A higher threshold of lexical overlap between the context sentence and the question can increase the inference difficulty on the dataset because a higher ROUGE-1 score implies that the context sentence and the question either have a similar structure or describe the same thing, making it challenging to determine whether the sentence entails the answer to the question or is merely related to the question. In contrast, a low ROUGE-1 score indicates that the context sentence and the question are not related at all, and therefore, not entailment. It is worth noting that sentence-question pairs with labeled entailment have inevitably higher ROUGE scores; therefore, all samples with entailment labels are pre-included, and the threshold is only applied to non-entailment samples. We evaluate the ChatGPT model with a 3-shot setting on the test set generated with different lexical overlap thresholds. The results in Table~\ref{Difficulty} show that inference accuracy increases with lower lexical overlap thresholds, indicating that by modifying the threshold, we can easily control the difficulty level of the generated dataset, which can be useful for assessing different parameter scale LLMs on a reasonably challenging dataset.


\begin{table}[!h]
\centering
\caption{We adopted ROUGE scores to set different levels of inference difficulty and assessed ChatGPT's performance on tasks of varying levels of difficulty.}
\label{Difficulty}
{
\begin{tabular}{lccccc}
\toprule

\multirow{2}{*}{Sample} & \multicolumn{2}{c}{Class distribution in test set} & \multirow{2}{*}{ACC} \\
\cmidrule(lr){2-3}
 & entail & not entail &\\
\midrule
R1$\geq$0.3 & 460&193 & 0.6186±0.0054\\
R1$\geq$0.2 & 460&517 & 0.6986±0.0066\\
R1$\geq$0.1 & 460&1327 & 0.7934±0.0044\\
R1$>$0 & 460&1681 & 0.8100±0.0023 \\
\bottomrule
\end{tabular}
}
\end{table}

\section{Discussion and Conclusion}

\subsection{Domain-Specific NLI Datasets}
Large-scale datasets constructed by crowdworkers have greatly contributed to the progress of domain-agnostic NLI. However, such datasets are prone to statistical artifacts that arise during the annotation process, as previous studies have shown \cite{poliak2018hypothesis,gururangan2018annotation,tsuchiya2018performance}. These artifacts pose a significant challenge in the development and evaluation of NLI models, as they can lead to hypothesis-only classifiers achieving better-than-random performance. To address this challenge, some researchers have attempted to create more robust and challenging datasets \cite{nie2019adversarial}. In this study, we propose a new dataset, RadQNLI, specifically designed for the radiology domain, with the help of multiple experts. However, developing a highly specialized dataset requires professional expertise, which makes it difficult to generate large-scale data in a short period. Therefore, our dataset has a smaller volume of data compared to open-domain datasets. To ensure a comprehensive evaluation of the dataset, we conducted a detailed statistical analysis and used ROUGE score to set up tasks with varying levels of inference difficulty during the experiments, in order to filter out possible statistical artifacts as much as possible. In future research, it will remain a crucial issue to incorporate and leverage the knowledge of experts to guide the automation of specialized dataset generation.

\subsection{Uniform vs Domain-Specific Models}
\begin{table}
\centering
\caption{Monetary costs (\$) and inference times (s/item) of ChatGPT and GPT4 on RadQNLI testset.}
\label{cost}
\begin{tabular}{lcccccc}
\toprule
 \multirow{2}{*}{Method} & \multicolumn{2}{c}{Length (tokens)} & \multicolumn{2}{c}{ChatGPT} & \multicolumn{2}{c}{GPT4}  \\
 \cmidrule(lr){2-3} \cmidrule(lr){4-5} \cmidrule(lr){6-7}
 & prompt & completion & cost & time & cost & time\\ 
\midrule
Zero-shot+IRSA & 130.86 & 2.63 & 0.26 & 0.41 & 3.99 & 0.90\\
Zero-shot+CoT & 144 .86 & 66.13 & 0.41 & 2.40  & 8.12 & 12.45 \\
Few-shot+IRSA & 605.86 & 1.54 & 1.19 & 0.43 & 17.85 & 0.96\\
Few-shot+CoT & 1007.86 & 41.77 & 2.05 & 1.62 & 31.99 & 5.41\\
\bottomrule
\end{tabular}
\end{table}
Creating a unified, general-purpose model to solve problems in all fields versus creating local specific models for different fields is an important issue. Especially at this particular point in time, where LLMs represented by ChatGPT and GPT-4 have achieved unprecedented success in different tasks, bringing new hope for creating a unified model. In this study, we aimed to explore this issue further by comparing the performance of ChatGPT/GPT-4 and locally fine-tuned models in a highly specialized task - radiology language inference. The results indicated that local models' performance falls short of that of ChatGPT/GPT-4 when the amount of specialized data is limited. However, with sufficient data, local models outperform the unified large models. Therefore, we are faced with the dilemma of choosing between building a unified, general-purpose model or creating effective datasets, both of which require substantial resources when reaching a certain scale (details can be seen in Table~\ref{cost}). Alternatively, a more promising approach to this issue is to combine existing unified LLMs with limited specific datasets to better solve tasks in specialized domains.

In conclusion, in this work, we evaluated the performance of ChatGPT/GPT-4 on a radiology NLI task and compared it to other models fine-tuned specifically on task-related data samples. The results showed that GPT-4 outperformed ChatGPT in the radiology NLI task, and other specifically fine-tuned models require significant amounts of data samples to achieve comparable performance to ChatGPT/GPT-4. These findings demonstrate that constructing a unified model capable of solving various tasks across different domains is feasible. However, further research is needed to determine whether this approach is preferable to creating domain-specific models, and how to effectively utilize limited specialized datasets with large LLMs.


\clearpage
\makeatletter
\renewcommand{\@biblabel}[1]{\hfill #1.}
\makeatother

\end{sloppypar}
\bibliographystyle{vancouver}
\bibliography{mybib} 
\end{document}